%% file: main.tex
\pdfoutput=1

\documentclass[11pt]{article}

\usepackage{acl}

\usepackage{times}
\usepackage{latexsym}

\usepackage[T1]{fontenc}

\usepackage[utf8]{inputenc}

\usepackage{microtype}
\usepackage{inconsolata}
\usepackage{subfig}
\usepackage{graphicx} 
\usepackage{multirow}
\usepackage{float}
\usepackage{amsmath}
\usepackage{microtype}
\usepackage{booktabs}
\usepackage{xspace}
\usepackage{stfloats}

\usepackage{xcolor}

\newcommand{\Method}{\textit{MODABS}\xspace}
\newcommand{\Cluster}{\textit{Cluster}}
\newcommand{\Keywords}{\textit{Keywords}}
\newcommand{\Prompting}{\textit{Prompting}}
\newcommand{\AspDiff}{\#AbsAspDiff}
\newcommand{\DDABS}{Disordered-DABS\xspace}
\newcommand{\DCnnDM}{D-CnnDM\xspace}
\newcommand{\DWikiHow}{D-WikiHow\xspace}
\newcommand{\OASUM}{OASUM\xspace}

\title{\Method{}: Multi-Objective Learning for Dynamic Aspect-Based Summarization}

\author{Xiaobo Guo \and Soroush Vosoughi\\
        Department of Computer Science \\ Dartmouth College \\
    Hanover, New Hampshire\\
    \{xiaobo.guo.gr, soroush.vosoughi\}@dartmouth.edu
    }
\begin{document}
\maketitle
\begin{abstract}

The rapid proliferation of online content necessitates effective summarization methods, among which dynamic aspect-based summarization stands out. Unlike its traditional counterpart, which assumes a fixed set of known aspects, this approach adapts to the varied aspects of the input text. We introduce a novel multi-objective learning framework employing a Longformer-Encoder-Decoder for this task. The framework optimizes aspect number prediction, minimizes disparity between generated and reference summaries for each aspect, and maximizes dissimilarity across aspect-specific summaries. Extensive experiments show our method significantly outperforms baselines on three diverse datasets, largely due to the effective alignment of generated and reference aspect counts without sacrificing single-aspect summarization quality.

\end{abstract}
\section{Introduction}
\begin{figure}[!hbt]
    \centering
    \includegraphics[width=0.99\columnwidth]{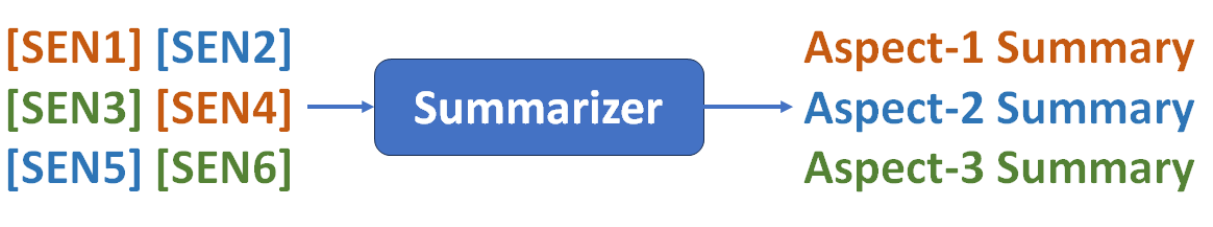}
    \caption{Diagram illustrating aspect-based summarization, with distinct colors representing different aspects. "[SEN $i$]" indicates the $i$-th sentence in the source article.}\label{fig:aspect-summarization}
    \end{figure}

With the rapid increase in digital content, the need for automated text summarization systems has grown significantly, as summaries efficiently distill key information from extensive texts. Among various approaches, query-focused summarization~\cite{wang2022squality,zhong2021qmsum,zhu2022transforming}  and aspect-based summarization~\cite{hayashi2021wikiasp,ahuja2022aspectnews} are particularly prominent for generating content-specific summaries. These methods, referenced in recent studies, cater to diverse information needs by emphasizing particular aspects or queries within the summaries. Aspect-based summarization, illustrated in Figure\ref{fig:aspect-summarization}, specifically focuses on extracting and summarizing information relevant to predefined aspects, offering a targeted approach to understanding large datasets.

    However, both methodologies encounter substantial constraints with the requirement of pre-specified aspects, thereby limiting their applicability. Dynamic aspect-based summarization, a recently proposed task~\cite{tan2020summarizing, maddela2022entsum}, requires models to automatically identify and extract relevant aspects from a given text, removing the need for prior knowledge about aspect numbers or content. Prior research on DABS typically adopts a two-phase methodology, initially focusing on generating aspect-based elements—such as identifying key sentences~\cite{hayashi2021wikiasp,amar-etal-2023-openasp} or keywords~\cite{souza2022aspect,yang-etal-2023-oasum,zhu2021twag} relevant to each aspect. This step is followed by employing a conventional summarization model that integrates the identified aspect information with the main article. An alternative approach utilizes the capabilities of large language models (LLMs), such as GPT-3.5-turbo, for aspect-based summarization without predetermined aspects~\cite{yang2023exploring, Guo2024DDABS}. However, despite LLMs' potential, their effectiveness in DABS is often lower than the two-step approach, mainly due to their insufficient explicit aspect knowledge and prior information~\cite{Guo2024DDABS}.

    
    To solve this task, we propose a novel \textbf{m}ulti-\textbf{o}bjective \textbf{d}ynamic \textbf{a}spect-\textbf{b}ased \textbf{s}ummarization framework, which we call \Method{}. Our framework dynamically identifies and summarizes the aspects without needing prior knowledge. Our multi-objective model not only works towards the conventional summarization objective but also predicts the number of aspects and works towards maximizing the divergence (i.e., minimizing the overlap) between the summaries of different aspects. Empirical evaluations on three datasets for this task show that our framework outperforms the existing models in predicting the number of aspects and generating high-quality aspect-based summaries. Furthermore, our ablation analysis reveals the crucial role of multi-objective learning in enhancing performance.

    The contributions\footnote{The code and data for this paper are available here: https://github.com/xiaobo-guo/MODABS} of this work are as follows: 
    \begin{itemize}        
        \item We propose a novel multi-objective framework, called \Method{}, that enables the generation of multiple aspect-specific summaries from a single input without requiring prior knowledge of the aspects. Alongside the standard summarization objective, we incorporate the divergence of different aspect-based summaries and the prediction of the aspect number to enhance the performance of our method. The versatility of our framework allows it to be incorporated with any encoder-decoder or decoder-only model.
        
        \item Through our ablation analysis, we demonstrate the effectiveness of these additional objectives in improving dynamic aspect-based summarization. By controlling the distance between the generated summaries and the precision of the aspect numbers, our approach achieves enhanced performance.
    \end{itemize}

\section{Related Work}
    \subsection{Datasets for Aspect-Based Summarization}   
        Numerous studies have explored aspect-based summarization, predominantly in the context of reviews such as those for products or restaurants~\cite{lu2009rated, wang2016neural,yang2018aspect,angelidis2021extractive,amplayo2021aspect,chu2019meansum}. Recently, the research scope expanded to include news articles, which facilitated the creation of artificially engineered datasets with pre-defined aspects~\cite{frermann2019inducing, ahuja2022aspectnews}. Encyclopedia data, another substantial information source, has been leveraged for crafting aspect-based summarization datasets~\cite{hayashi2021wikiasp}. This diversification tests the generalizability of models across different datasets. The entity-based summarization dataset proposed by~\citet{maddela2022entsum} offers another angle on aspect-based summarization.
        
        Nevertheless, a significant limitation persists across these datasets: they all have pre-defined aspects. Consequently, the aspects related to a domain's samples remain static, simplifying the task considerably. Therefore, dynamic aspect-based summarization is proposed where aspects are neither limited nor predefined. This approach encompasses datasets that identify entities (e.g., AnyAspect~\cite{tan2020summarizing}, ENTSUMV2~\cite{maddela2022entsum}) or conceptual labels/sub-topics within the text (e.g., OASUM~\cite{yang-etal-2023-oasum}, OpenAsp~\cite{amar-etal-2023-openasp}). Furthermore, extensions like disordered-DABS~\cite{Guo2024DDABS} add complexity by introducing disordered aspect information, where sentences related to the same aspect may not appear in sequence.

    \subsection{Methods in Dynamic Aspect-Based Summarization}
        Existing methods for dynamic aspect-based summarization (DABS) have primarily evolved from traditional aspect-based summarization techniques, where models leverage automatically generated aspect information to produce summaries. The landscape of DABS is predominantly shaped by two frameworks: one that segments the task into sentence grouping and summarization, and another that generates summaries based on predetermined aspect keywords. 
        
        The first approach uses a two-step procedure to deliver aspect-specific summaries~\cite{liugenerating}. Angelidis and Lapata~\cite{angelidis2018summarizing} proposed a weakly supervised method for this task, leveraging a topic extractive component and a summarizer. These techniques typically group sentences using domain-specific information. For improved model generalization, recent studies have favored supervised methods for sentence grouping, utilizing trained classifiers~\cite{hayashi2021wikiasp,amplayo2021aspect}, or sentence embeddings similar to identified topic sentences~\cite{angelidis2021extractive}. 
        
        An alternative method for aspect-based summarization is the controllable generation, commonly applied across various domains such as politeness~\cite{sennrich2016controlling}, content~\cite{fan2018controllable}, and style~\cite{ficler2017controlling}. Based on the idea of controllable generation~\citet{he-etal-2022-ctrlsum} use keywords as prompts, which was further enhanced by~\citet{ahuja2022aspectnews}, who devised a supervised method to select aspect-oriented keywords. Note that when applying these methods to the DABS, the supervised method for sentence clustering and keyword generation shifts to an unsupervised approach, like BERTopic~\cite{grootendorst2022bertopic}, due to the lack of ground truth in the initial phase of aspect-based information extraction~\cite{yang-etal-2023-oasum,Guo2024DDABS,maddela2022entsum}.
        
        Beyond these conventional methods, an alternative one-step approach utilizes prompting techniques alongside LLMs, such as GPT-3.5-turbo or GPT-4~\cite{goyal2022news, yang2023exploring,Guo2024DDABS}. This method relies on crafted prompts to direct LLMs in generating aspect-based summaries from the given inputs. Given that LLMs require no prior domain-specific knowledge, they are theoretically well-suited for addressing the challenges of dynamic aspect settings. Despite this, the \Prompting{} method has shown limited success in such contexts, as identified by~\cite{Guo2024DDABS}, primarily due to the insufficiency of aspect information and prior knowledge, which significantly hinders the model's ability to accurately fulfill the task.
        
        To overcome these challenges, our study introduces a novel multi-objective approach that leverages weak-supervised techniques. This method allows language models to implicitly understand aspect-based information, aiding in the generation of coherent aspect-based summaries. By integrating weak supervision, our approach addresses the complexities of DABS more effectively, sidestepping the need for explicit aspect annotations or extensive preprocessing.

\section{Datasets}\label{sct:datasets}

    In our study, we focus on three datasets designed for dynamic aspect-based summarization: \DDABS{}~\cite{Guo2024DDABS} and \OASUM{}~\cite{amar-etal-2023-openasp}.

\DDABS{} is crafted for disordered dynamic aspect-based summarization, encompassing two datasets: \DCnnDM{}, adapted from CNN/Daily Mail~\cite{see2017get}, and \DWikiHow{} from WikiHow~\cite{koupaee2018wikihow}. These datasets utilize existing summarization datasets to generate aspect-based summaries by merging multiple samples (\DCnnDM{}) or segmenting topics into detailed steps (\DWikiHow{}), introducing shuffled input sentences to mimic the disorder often found in online discussions.

 \OASUM{} is a dynamic aspect-based summarization dataset containing over three million samples. Unlike \DDABS{}, aspects in \OASUM{} may exhibit hierarchical structures, where one aspect may be nested within another, e.g., ``History'' and ``History: Japanese invasion''. To ensure a fair comparison with other baselines, we preprocess \OASUM{} to retain only aspects with the narrowest scope. Similar to \DDABS{} and \citet{frermann2019inducing}, sentence shuffling is introduced to create disorderly scenarios.

Both DDABS and OASUM datasets include samples with very long inputs, single-aspect summaries, or multiple aspects. Following~\citet{Guo2024DDABS} and ~\citet{yang-etal-2023-oasum}, we preprocess these datasets by truncating source articles and summaries and limiting the number of aspects based on their average length and variability, as detailed in Appendix \ref{sct:Data_process}. Given that the original \OASUM{} dataset includes about half of its samples with only one aspect, we exclude these single-aspect samples from our experiments.

\section{Methodology}

\begin{figure*}
        \centering
        \includegraphics[width=1.99\columnwidth]{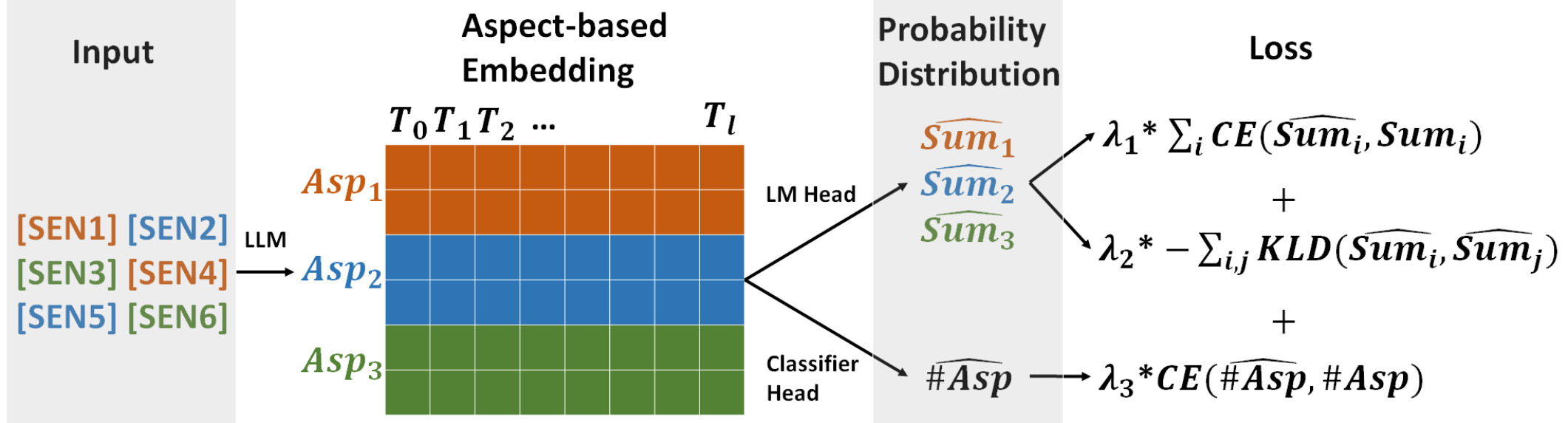}
        \caption{Diagram of our framework. Colored sentences represent different aspects. ``[SEN i]'' indicates the i-th sentence in the input. Aspects, tokens, and their generated summaries are denoted as \(Asp_i\), \(T_i\), and \(\widehat{Sum_i}\), respectively. The predicted number of aspects is \(\widehat{\#Asp}\), while ground-truth summaries and aspect numbers are \(Sum_i\) and \( \#Asp\), respectively. Cross-entropy loss and KL divergence loss are indicated by ``CE'' and ``KLD''. Weights for different losses are \(\lambda_{1/2/3}\).}\label{fig:model}
    \end{figure*}

    Our fine-tuning process, illustrated in Figure~\ref{fig:model}, diverges from standard summarization fine-tuning in two key areas: First, we modify the decoder embeddings with the intent to account for multiple aspects, a feature not commonly considered in traditional summarization tasks. By transforming the embeddings, our model generates multiple aspect-specific summaries concurrently, offering a more comprehensive summary of the input text. Second, we incorporate a multi-objective learning scheme into our framework. 
    
    These objectives include not only the standard text summarization task but also the maximization of the distance between different aspect-based summarization representations. Furthermore, our model is designed to predict the number of aspects based on the decoding representation. In effect, the multi-objective learning approach encourages our model to create more distinct and specific summaries per aspect, thus enhancing the granularity and precision of the summarization.

    \subsection{Decoder Embeddings}
        In traditional encoder-decoder or decoder-only models, the decoder generates embeddings (\(E\)) of shape \(B\times L\times D\), where \(B\) is the batch size, \(L\) is the summary length, and \(D\) is the dimensions of all heads. This embedding is then passed to a language model (LM) head to generate the token distribution. Our model, aiming to generate multiple aspect-specific summaries simultaneously, reshapes \(E\) into \(B\times L\times N\times D_n\), where \(N\) is the predefined maximum number of aspects and \(D=N\times D_n\). To produce \(N\) aspect-specific summaries (\(\widehat{Sum_1},\widehat{Sum_2},\cdots\widehat{Sum_N}\)), we apply an LM head to these aspect-based embeddings. Following standard convention, the model predicts the number of aspects using the embeddings of all aspect-based summaries' first tokens (\(E[:,:,0,:]\)). This transformed embedding is then sent to a classifier head to estimate the aspect number (\(\widehat{\#Asp}\)). 
    \subsection{Multi-Objective Learning}
    Our model incorporates two additional objectives beyond standard text summarization: maximizing the distance between aspect-based summarization representations and predicting the aspect count using the decoding representation.
    Our loss function, therefore, comprises three components: (1) The cross-entropy (CE) loss between generated (\(\widehat{Sum_i}\)) and reference (\(Sum_i\)) summaries (Eq.~\ref{eq:summarization_loss}); (2) The KL divergence (KLD) between different aspect-specific summaries, bounded by a limit function (\(L\)) (Eq.~\ref{eq:KLD_loss}); (3) The cross-entropy loss between predicted (\(\widehat{\#Asp}\)) and reference (\( \#Asp\)) aspect numbers (Eq.~\ref{eq:aspect_loss}). The final loss for a model with a maximum aspect count of \(N\) is then computed as:
    \begin{align}
        Loss = &\lambda_1*\frac{\sum_i{CE(\widehat{Sum_i},Sum_i)}}{N}  \label{eq:summarization_loss} \\
                +&\lambda_2 * -\frac{\sum_{i,j}L(KLD(\widehat{Sum_i}, \widehat{Sum_j}))}{N}  \label{eq:KLD_loss} \\
                +&\lambda_3 * CE(\widehat{\#Asp}, \#Asp) \label{eq:aspect_loss}
    \end{align}
    \(\lambda_1\), \(\lambda_2\), and \(\lambda_3\) are weights assigned to each objective. The limit function, \(L\), can be either a Sigmoid or Tanh. More details on the weighting of the loss and the limit function are given in Appendix~\ref{sct:weight_and_limitfunc}. The intuition behind the multi-objective setting is expanded on in Appendix~\ref{sct:Multi-objective-intution}.

        \begin{table*}[!htb]
        \centering
        \setlength\tabcolsep{3pt}
        \begin{tabular}{lllllll}
        \hline
        Dataset   & Model         & \AspDiff{}      & BERTScore   & Rouge-1     & Rouge-2    & Rouge-L     \\ \hline
        D-CnnDM   & Keywords      & 1.3 (0.0)*      & 14.2 (0.1)* & 24.8 (0.0)* & 8.9 (0.2)* & 17.2 (0.1)* \\
                  & Cluster       & 1.3 (0.0)*      & 15.1 (0.2)* & 25.4 (0.1)  & 9.1 (0.1)* & 17.1 (0.1)* \\
                  & GPT-3.5-Zero  & 6.2            & 9.1        & 12.4        & 4.1       & 9.1         \\
                  & GPT-3.5-Tuned & 3.5             & 12.9       & 18.4        & 6.7       & 11.9       \\
                  & \Method{}        & \textbf{1.0 (0.1)}       & \textbf{18.2 (0.4)}  & \textbf{26.4 (0.5)}  & \textbf{9.7 (0.2)}  & \textbf{18.3 (0.4)}  \\ \hline
        D-WikiHow & Keywords      & 2.7 (0.1)*      & 30.5 (0.1)* & 14.6 (0.2)* & 5.0 (0.1)* & 14.2 (0.2)* \\
                  & Cluster       & 2.7 (0.1)*      & 31.3 (0.0)* & 19.1 (0.1)* & 7.8 (0.1)* & 18.5 (0.1)* \\
                  & GPT-3.5-Zero  & 5.5             & 17.7        & 11.4        & 3.8        & 10.3        \\
                  & GPT-3.5-Tuned & 3.8             & 20.7        & 13.4      & 5.2        & 12.6        \\
                  & \Method        & \textbf{1.5 (0.1)}       & \textbf{40.6 (0.6)}  & \textbf{23.0 (1.0)}  & \textbf{9.5 (0.6)}  & \textbf{22.4 (0.9)}  \\ \hline
        OASUM     & Keywords      &1.5 (0.0)*     &21.8(0.8)*  &22.3 (0.1)* &11.0 (0.0)* &19.4 (0.0)*             \\
                  & Cluster       &1.5 (0.0)*     &20.7(0.1)* &21.2 (0.1)*  &10.0 (0.0)* &18.6 (0.00)  \\
                  & GPT-3.5-Zero  &4.2             &9.9          &11.9          &4.4           &9.0             \\
                  & GPT-3.5-Tuned &1.0             &14.2         &19.8          &8.9           &15.8             \\
                  & \Method        &\textbf{0.5 (0.0)}      &\textbf{28.5 (0.1)}   &\textbf{29.9 (0.1)}   &\textbf{14.9 (0.1)} &\textbf{26.0 (0.1)}  \\ \hline
        \end{tabular}
            \caption{The performance of our models and baselines across all three datasets. Mean scores are reported, accompanied by standard deviations in brackets. An asterisk (*) indicates a statistically significant difference (p<0.05) between the baseline models and our \Method{} implementation. Due to budgetary constraints, the results for GPT-3.5-Zero and GPT-3.5-Tuned are derived from a single experimental run; consequently, standard deviations and p-values could not be computed for this model.}\label{tab:summary_results}
        \end{table*}

\section[Experiments]{Experiments\footnote{The computing infrastructure is detailed in Appendix~\ref{sct:computing_infrastructure}}}
    \subsection{Baselines and Metrics}

     We compare our model against three baselines consistent with those used in the \DDABS{} and OASUM datasets:  \Cluster{}, \Keywords{}, and \Prompting{}.
     
    For \DDABS{} (specifically, the \DCnnDM{} and \DWikiHow{} sub-datasets), our benchmarks align with those reported by Guo et al.~\cite{Guo2024DDABS} for both the \Cluster{} and \Keywords{} strategies. For \OASUM{}, we adhere to the aspect-based summarization methodology as detailed by Guo et al.\cite{Guo2024DDABS}. Notably, for the \Keywords{} method, we opt for a hyper-parameters-tuned BERTopic model for keyword generation, diverging from the utilization of Wikipedia subtitles to mitigate potential performance inflation due to leaked aspect information.
    
    We extend our exploration of the \Prompting{} strategy to encompass both zero-shot learning outcomes and few-shot fine-tuning results, as reported by Guo et al.~\cite{Guo2024DDABS}, leveraging the latest OpenAI API capabilities for fine-tuning the GPT-3.5-turbo model to accept up to 16k input tokens. Detailed insights into GPT-3.5-turbo are provided in Appendix \ref{sct:GPT-3.5-turbo-details}. The prompts are designed as follows: We generated five initial prompts tailored to the summarization task and selected the most effective one based on performance. This prompt was then refined through multiple iterations,  to ensure adherence to format, aspect, and length requirements. The finalized prompt emphasizes the generation of concise, aspect-specific summaries.
    
    Our evaluation employs a suite of automatic metrics: BERTScore, Rouge-1, Rouge-2, Rouge-L, and \AspDiff{}. These metrics, assessing summary quality through precision, recall, and linguistic quality, are complemented by \AspDiff{}, which specifically quantifies the discrepancy in aspect numbers between reference and generated summaries, employing the aspect alignment methodology from Guo et al.~\cite{Guo2024DDABS} for a consistent comparison framework. We also control the generated summary because of its influence of the performance shown in ~\citet{guo-vosoughi-2023-length}
    
    In addition to automatic metrics, we incorporate human annotation scores, evaluating summaries on ``Coherence'', ``Consistency'', ``Fluency'', ``Relevance'', ``Aspect Quality'', and ``Overall Rank'', as informed by previous work~\cite{fabbri2021summeval,Guo2024DDABS,amar-etal-2023-openasp,yang-etal-2023-oasum}. These human-centric scores ensure a comprehensive assessment of summary quality from both technical and perceptual standpoints. The criteria and details pertaining to human annotators are elucidated in Appendix \ref{sct:annotators}.

    \subsection{Automatic Evaluation Results}\label{sct:results}
        The performance of our \Method model, alongside baseline methods, is summarized in Table~\ref{tab:summary_results}. These results leverage the Longformer-Encoder-Decoder~\cite{beltagy2020longformer} used by both datasets for its ability to take more than 10k input tokens.

    As shown in Table~\ref{tab:summary_results}, \Method excels in all evaluation metrics across the three datasets. The performance improvements over baseline models are statistically significant (p<0.05). The smallest performance gains are seen on the \DCnnDM{} dataset, whereas \OASUM{} exhibits the most marked improvements.
    
    This suggests that our model's efficacy may be tied to the intricacy of distinguishing between aspects. Given that \DCnnDM{} aggregates multiple news articles, it inherently presents fewer complexities, accounting for the smaller gains observed. GPT-3.5-Zero performs sub-optimally due to its deployment as an off-the-shelf model with zero-shot inference. In line with the findings reported by~\citet{Guo2024DDABS}, our study observes that the performance of GPT-3.5-Tuned, when applied to the DABS task, significantly exceeds that of zero-shot learning approaches. 
    
    This enhancement underscores the value of fine-tuning in tailoring model responses to the specific nuances of DABS tasks. Despite this improvement, it is noteworthy that GPT-3.5-Tuned does not reach the performance benchmarks set by the two-step baseline models or our proposed \Method{}. This underscores the advanced capability of \Method{} in dynamically navigating the challenges of aspect-based summarization, further reinforcing the necessity for specialized approaches in handling the intricacies of such tasks.

    Analyzing the absolute aspect count differences between reference and generated summaries, \Method{} exhibits consistent performance across datasets, unlike the variable outcomes from other baselines. This stability implies that, in addition to aspect-based summarization, our model effectively predicts aspect numbers. We validate this aspect number prediction against two baseline methods: (1) a fine-tuned classifier and (2) the BERTopic model with adjusted hyperparameters. 
    
    \begin{table}[!htb]
    \small
        \centering
        \begin{tabular}{llll}
        \toprule
                    & \DCnnDM{}            & \DWikiHow{}           & \OASUM{}   \\ \midrule
        Classifier  & 1.69 (0.34)          & 1.89 (0.02)          & \textbf{0.48 (0.00)}           \\
        Cluster     & 1.28 (0.00)          & 2.69 (0.06)          & 1.48 (0.00) \\
        \Method{}   & \textbf{1.04 (0.01)} & \textbf{1.48 (0.07)} & 0.51 (0.01) \\ \bottomrule
        \end{tabular}
        
        \caption{Absolute aspect number difference between model predictions and ground-truth summaries. ``Classifier'' and ``Cluster'' correspond to predicting the number of aspects using a classifier and BERTopic, respectively.}\label{tab:aspect_number}
    \end{table}
    
    Table~\ref{tab:aspect_number} demonstrates \Method superiority over the ``Cluster'' method across all datasets. Compared with the ``Classifier'' method, \Method{} performs better on \DCnnDM{} and \DWikiHow{} and almost the same on \OASUM{}. This unexpected result underscores the benefit of incorporating the aspect number prediction objective into our model's loss function.

        \begin{table*}[!hbp]
        \centering
        \setlength\tabcolsep{5pt}
        \begin{tabular}{lcccccc}
        \toprule
            Model    & Coherence   & Consistency & Fluency     & Relevance   & Aspect Quality & Rank        \\
            \midrule
            GPT-3.5-Zero  & 3.21 (1.28) & 2.71 (0.75) & 4.29 (1.12) & 3.12 (0.80) & 4.25 (1.07)    & 2.04 (0.95) \\
            Keywords & 3.17 (0.76) & 2.79 (0.93) & 3.38 (0.97) & 3.25 (0.68) & 3.00 (1.14)    & 3.12 (0.99) \\
            Cluster  & 3.50 (0.72) & 3.21 (0.78) & 3.50 (0.72) & 3.46 (0.88) & 4.04 (1.00)    & 2.62 (0.71) \\
            \Method      & \textbf{3.83 (0.48)} & \textbf{3.79 (0.66)} & \textbf{4.54 (0.78)} &\textbf{ 4.38 (0.77)} & \textbf{4.67 (0.64)}    & \textbf{1.21 (0.59)} \\
            \bottomrule
            \end{tabular}
            \caption{Average (std) human evaluation ratings (1--5 scale) on the five quality criteria and the ranking (1--4 scale), determined by 30 instances from the test samples. The results of GPT-3.5-Zero, \Keywords{}, and \Cluster{} except for the Rank is from~\citet{Guo2024DDABS}}\label{tab:results_human}
        \end{table*}

    \subsection{Human Evaluations}

    Human evaluation metrics provide a reference-free perspective, essential for assessing the quality of datasets and baseline models in the context of the source article. These evaluations concentrate on two primary aspects: the overall quality of the summarization and the quality of the aspects captured.
    
    Following \citet{fabbri2021summeval}, we assess summaries using the following criteria: (1) \textit{Coherence}, assessing the structural integrity and logical flow of the summary; (2) \textit{Consistency}, ensuring the summary's factual content aligns with the source, particularly by avoiding the inclusion of information not present in the source articles; (3) \textit{Relevance}, highlighting the inclusion of critical content from the source; and (4) \textit{Fluency}, scrutinizing the grammatical and linguistic quality of the summary to guarantee readability.

    In addition to these criteria, the unique requirements of DABS necessitate an extra metric, \textit{Aspect-Quality}. This metric is crucial for evaluating the effectiveness with which aspect-based summaries maintain distinctiveness and focus. It ensures that each aspect is not only delineated but also remains the focal point within its designated summary~\cite{angelidis2021extractive,amplayo2021aspect,Guo2024DDABS}.

    The human evaluations were conducted on samples from the \DCnnDM{} dataset, following the methodology and insights from~\citet{Guo2024DDABS}. Annotators compared summaries from our method against baselines using the five criteria above and ranked them to determine overall quality. Details about the annotators and their compensation can be found in Appendix~\ref{sct:annotators}.

    The results, detailed in Table \ref{tab:results_human}, show our method outperforming baselines across all criteria. We divided evaluation criteria into single-aspect quality (Coherence and Fluency) and multi-aspect quality (Consistency, Relevance, and Aspect Quality), with our method showing marked improvements, especially in multi-aspect quality. This suggests our multi-objective learning framework effectively enhances the model's ability to identify and synthesize relevant aspect-based information, particularly in handling complex summarization tasks.

\section{Ablation Analysis}
  In this section, we present examples of summaries generated using different methods. We also conduct ablation analyses to investigate (1) The influence of the multi-objective settings and  (2) The influence of discrepancies in the aspect numbers between the reference and generated summaries.

    \subsection{Examples of Summaries}
        \begin{figure*}[!hbt]
            \centering
            \includegraphics[width=1.99\columnwidth]{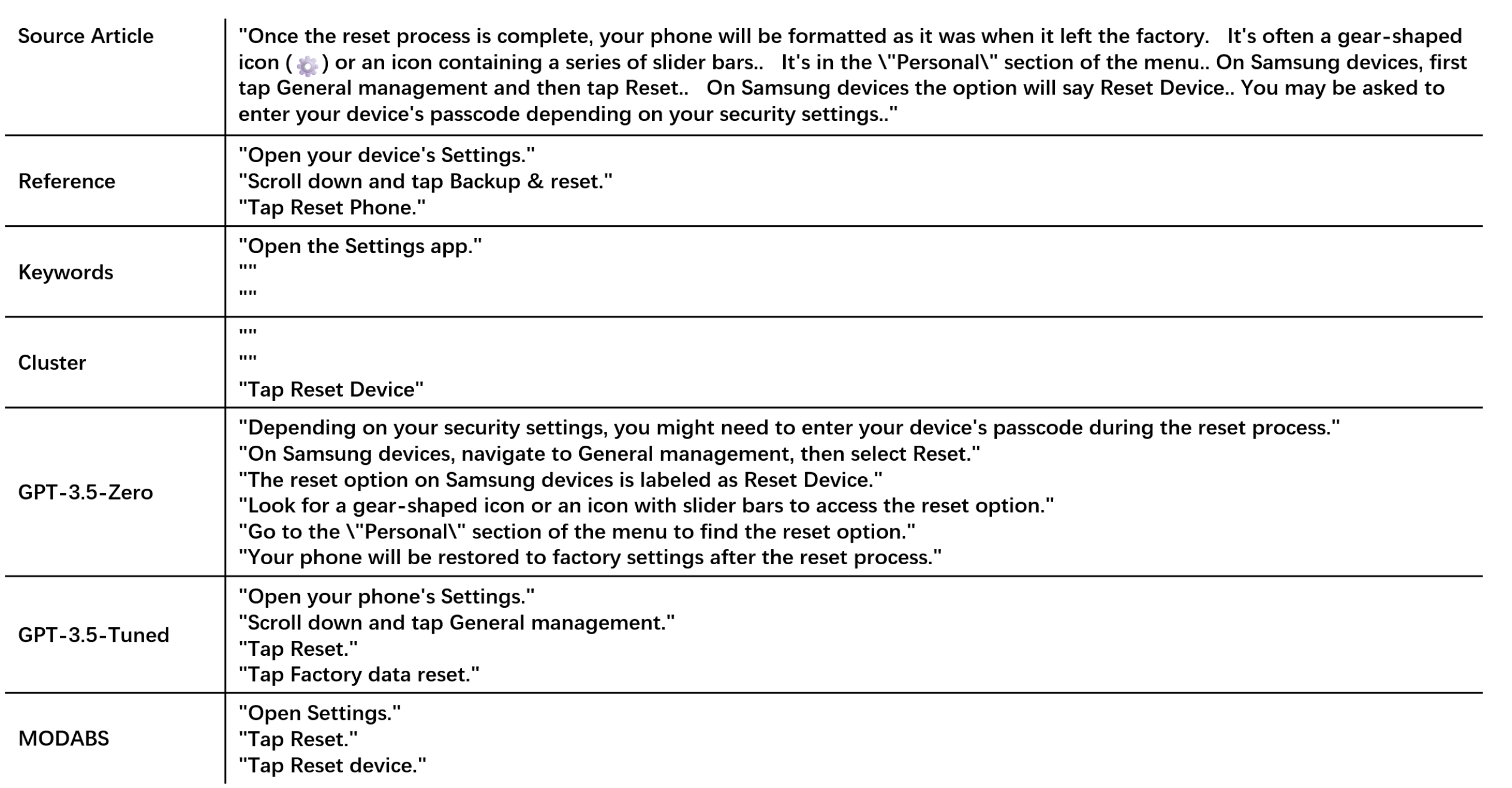}
            \caption{An example of the source article, reference, and the generated Summaries. Empty quotes (``\,'') indicate that no generated summaries correspond to this reference summary.}\label{fig:summary_sample}
        \end{figure*}
        
        Figure~\ref{fig:summary_sample} offers an illustrative example that includes the source article, reference summaries, and summaries generated by both the baseline models and \Method{}. The source article discusses the procedure for resetting a phone to its factory settings. For easier comparison, we have aligned the references and generated summaries. Instances of empty quotes (``\,'') indicate the absence of a corresponding generated summary for a given reference summary. Notably, \Method{} produces two aspect-based summaries, highlighting its capability to address multiple aspects simultaneously. In contrast, the \Keywords{} and \Cluster{} methods generate summaries focused on a single aspect only.
        
        Regarding the \Prompting{} method, the summary generated by GPT-3.5-Zero is notably as long or even longer than the source article itself, indicating inefficiency in condensing information. GPT-3.5-Tuned shows an attempt to cover all three identified aspects but introduces an additional aspect not present in the source article, demonstrating a tendency towards hallucination. This issue is critical in summarization tasks, as it can compromise the accuracy and reliability of the generated summaries.

    \subsection{Influence of the Multiple Objectives}
    To assess the impact of various multi-objective configurations on model performance, we analyzed a 20\% sample of the dataset. Detailed settings of the loss weights for these experiments are provided in Appendix \ref{sct:weight_and_limitfunc}. The outcomes of these configurations are summarized in Table \ref{tab:multitask_analysis}, showcasing the effects of different objective combinations across all three datasets.

    \begin{table*}[!ht]
            \centering
            \begin{tabular}{ll|lllll}
            \hline
            Dataset                   & Model & \AspDiff{}       & BERTScore      & Rouge-1        & Rouge-2        & Rouge-L        \\
            \multirow{3}{*}{\DCnnDM{}}   & +Asp  & 1.27          & 10.39          & 20.38          & 6.17           & 14.51          \\
                                      & +KLD  & 1.30          & 12.71          & 21.82          & 7.23           & 15.37          \\
                                      & All   & \textbf{0.93} & \textbf{14.89} & \textbf{24.05} & \textbf{8.02}  & \textbf{17.00} \\ \hline
            \multirow{3}{*}{\DWikiHow{}} & +Asp  & 1.89          & 33.99          & 15.34          & 4.38           & 14.93          \\
                                      & +KLD  & 2.14          & 34.28          & 16.19          & 5.05           & 15.75          \\
                                      & All   & \textbf{1.63} & \textbf{38.37} & \textbf{21.91} & \textbf{8.57}  & \textbf{21.21} \\ \hline
            \multirow{3}{*}{\OASUM{}}    & +Asp  & 0.54          & 23.74          & 25.37          & 10.78          & 21.90          \\
                                      & +KLD  & 0.55          & 24.49          & 26.51          & 11.95          & 22.95          \\
                                      & All   & \textbf{0.52} & \textbf{24.80} & \textbf{26.73} & \textbf{12.03} & \textbf{23.19} \\ \hline
            \end{tabular}
            \caption{Optimal performance metrics for models employing various objectives. The terms ``+Asp'' and ``+KLD'' indicate the inclusion of additional loss metrics alongside the standard summary cross-entropy loss, where ``+Asp'' refers to the loss associated with predicting the number of aspects, and ``+KLD'' refers to the KL Divergence loss calculated between different aspect-specific summaries. The designation ``All'' represents the aggregation of all considered loss metrics.}
            \label{tab:multitask_analysis}
        \end{table*}

    The addition of Aspect number prediction loss (+Asp) directly enhances the model's proficiency in accurately predicting the number of aspects within summaries, as evidenced by improvements in the \AspDiff{} metric. This direct targeting, however, does not extend to improving the diversity among aspect-based summaries. Consequently, while \AspDiff{} benefits from this focus, other performance metrics are the worst among all three model architectures
    
    In contrast, the integration of KL Divergence loss (+KLD) demonstrates a broader impact on performance metrics, with notable exceptions. Although KLD generally fosters improvements across most metrics by enhancing differentiation between aspect-based summaries, it does not benefit the \AspDiff{} metric. This discrepancy stems from KLD's propensity to penalize mismatches in aspect-based summary distributions, which can inadvertently raise the KL Divergence in cases where an aspect-based summary is inaccurately identified as empty. It is important to note that the penalization for incorrect aspect number predictions inherently affects these scores, suggesting that the actual performance impact of KLD could be more favorable than initially apparent when solely considering its effect on aspect number accuracy.

    The true potential of our approach is unlocked through the combination of both Aspect number prediction and KL Divergence losses. This combination not only rectifies the individual limitations posed by each loss when applied in isolation but also capitalizes on their strengths to foster a comprehensive enhancement of model performance. By accurately predicting aspect numbers and promoting diversity in aspect-based summaries, the model achieves a significant, balanced improvement across all evaluated metrics. This synergy demonstrates the effectiveness of our multi-objective framework.

    \begin{figure*} [!hbt]
        \centering
        \subfloat[\DCnnDM{}]{
        \includegraphics[width=0.67\columnwidth]{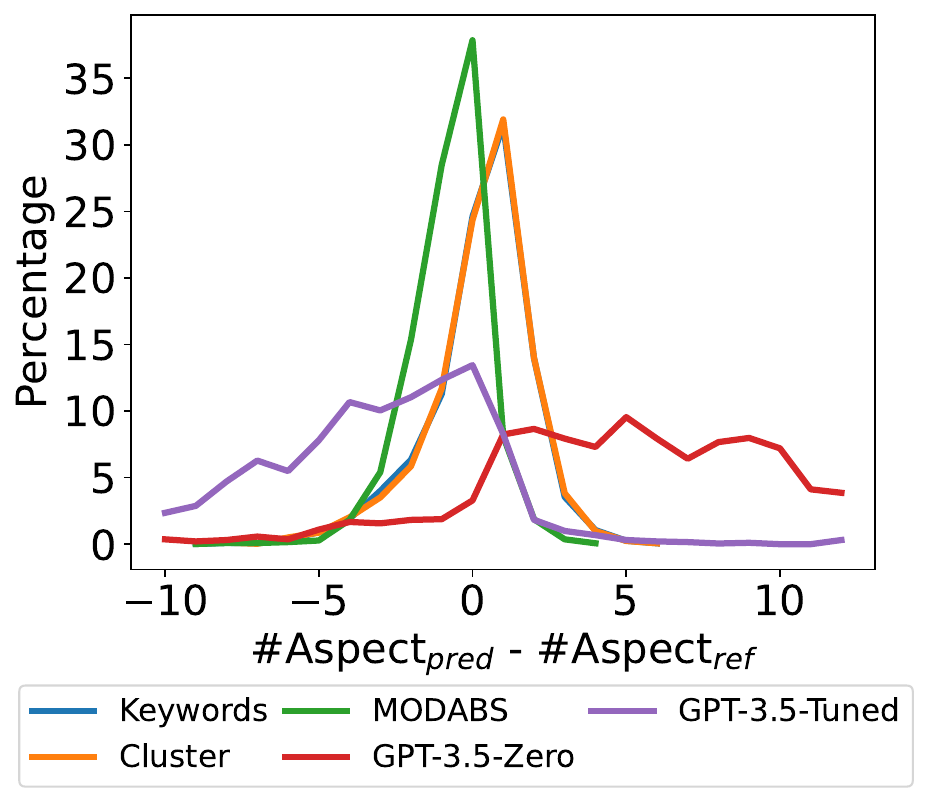}
        }
        \subfloat[\DWikiHow{}]{
        \includegraphics[width=0.67\columnwidth]{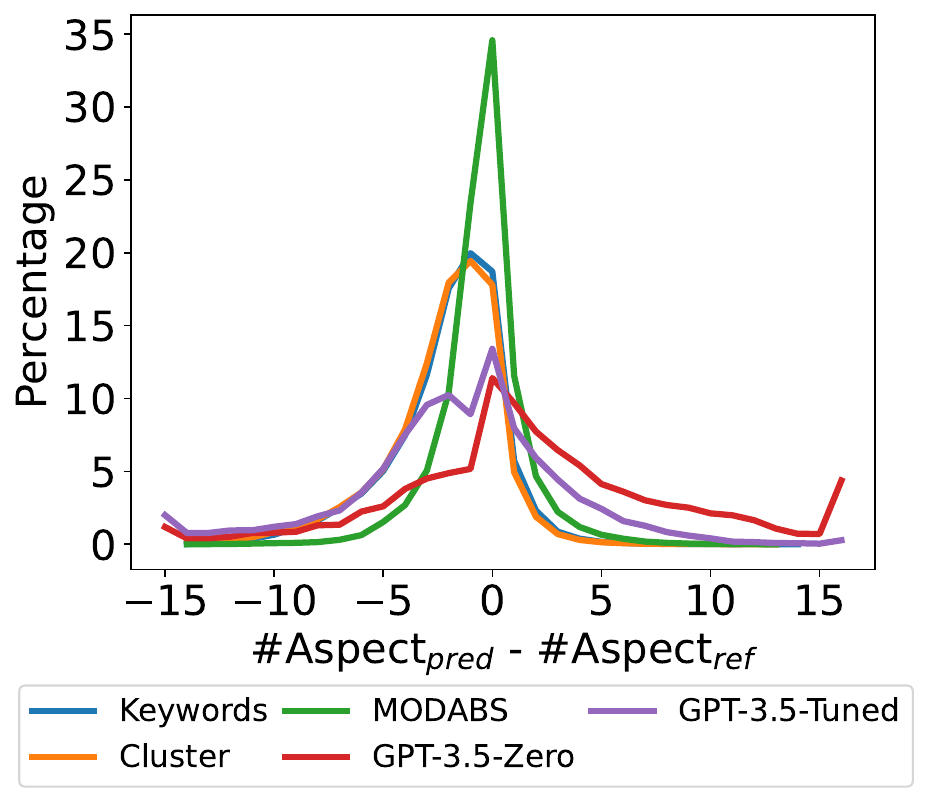}
        }
        \subfloat[\OASUM{}]{
        \includegraphics[width=0.67\columnwidth]{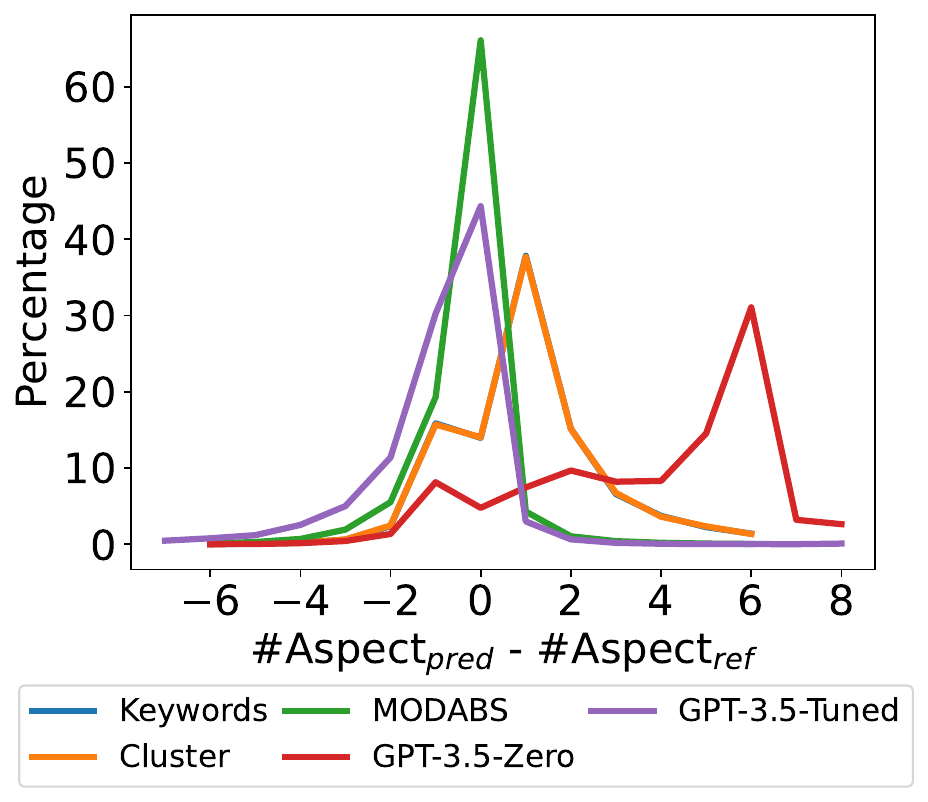}
        }
        \caption{Distribution of aspect number differences between reference and generated summaries for all three datasets.}\label{fig:aspdiff_distribution}
    \end{figure*}
    
    \subsection{Analysis of Aspect Number Discrepancies}
        The difference in aspect numbers between reference and generated summaries plays a significant role in the performance of the models of the DABS task. Hence, we examine the distribution of the aspect number discrepancies between references and generated summaries for the \Method{} and all baselines in Figure~\ref{fig:aspdiff_distribution}.

        As seen in Figure~\ref{fig:aspdiff_distribution}, \Method{} is more accurate at predicting the number of aspects compared to the baselines, which aligns with expectations since \Method{}'s learning objective includes an aspect number prediction component. This improvement is manifested in two distinct ways. Firstly, the pronounced peak at a ``0'' difference indicates that \Method{} more frequently aligns with the reference regarding aspect count, underscoring its precision in aspect prediction. Secondly, unlike other models, particularly GPT-3.5-Zero, \Method{} exhibits a zero-centric distribution. This pattern suggests a more refined comprehension of the dataset-specific aspect definitions by \Method{} and highlights its methodological advantage in understanding the nuanced structure of aspect-based information.
        
        The distributions associated with \Prompting{} methods, represented by GPT-3.5-Zero and GPT-3.5-Tuned, further illuminate the dynamics of aspect prediction. GPT-3.5-Zero tends to overestimate the number of aspects, as evidenced by a right-shifted curve, whereas fine-tuning through GPT-3.5-Tuned corrects this bias, resulting in a distribution that is slightly left-shifted, indicating fewer aspects predicted than the reference. This shift underscores the adaptability of LLMs to capture the definitional nuances of aspects through minimal tuning, with less than 200 samples required for effective learning. Interestingly, the BERTopic method, even without hyperparameter optimization, demonstrates an ability to surpass the performance of LLMs in this context. This observation may point to limitations within LLMs' capacity to grasp and replicate aspect-based summaries' complex, multifaceted nature without extensive customization or domain-specific tuning.

\section{Conclusion and Future Directions}

This study introduces a robust multi-objective framework tailored for dynamic aspect-based summarization, addressing the complexities of disordered information prevalent in today's digital landscape. By innovatively merging aspect discovery with aspect-based summarization, our approach surpasses traditional methods in accurately predicting aspect counts and significantly enhances summary quality. The empirical evidence drawn from extensive experiments across various datasets underscores the efficacy of our framework. 

Looking ahead, the integration of advanced attention mechanisms and adaptive layers promises to further refine our model, paving the way for even more precise and efficient summarization techniques.

\section{Limitations \& Ethial Considerations}
    We foresee no significant ethical issues related to our work. However, we employ three publicly accessible datasets that may contain sensitive or offensive material. Researchers are advised to approach these datasets with caution.

    When applying the findings and the method proposed in our study to summarization tasks, the following potential limitations should be taken into account:

    \textbf{Dependence on Loss Weights:} Our multi-objective framework's performance depends on the weights assigned to various objectives, which may differ from one dataset to another. Currently, we determine the optimal weight combination through small-scale dataset experiments, a process that is both time-consuming and not necessarily successful in finding the best combination. Future work could involve treating these weights as learnable parameters.
    
    \textbf{Model Compatibility: } In our research, we restricted our experimentation to the Longformer-Encoder-Decoder due to length constraints. Even though our framework is designed to be compatible with all encoder-decoder/decoder-only models, additional verification is required.
    
    \textbf{Memory Requirements:} The multi-objective learning process of DAS demands more GPU memory than the baseline models. Mitigating this constraint is a challenge, but potential solutions could involve the use of models like RWKV~\cite{peng2023rwkv}, or Hyena~\cite{poli2023hyena}, which have linear or non-dependent memory requirements relative to the input/output length.
    
    \textbf{Language Constraints: } Our experiments have been exclusively conducted in English. Although our method is theoretically language-independent, it is still uncertain whether our framework will effectively generalize to other languages. More tests are needed to affirm this.

\section*{Acknowledgments}
This work is supported in part by NSF Award 2242072 and a generous grant by the Templeton Foundation.

\bibliography{custom}

\setcounter{table}{0}
\setcounter{figure}{0}
\renewcommand\thefigure{\Alph{section}\arabic{figure}}
\renewcommand\thetable{\Alph{section}\arabic{table}}
\clearpage
\appendix

\input{appendix.tex}

\end{document}

%% file: appendix.tex
\section{Multi-objective Loss}\label{sct:Multi-objective-intution}
The parameters for our model's multi-objective framework are based on the following observations and considerations:

    \textbf{Cross-Entropy Loss:} For the summarization task, it is vital to calculate the cross-entropy loss between the generated and reference summaries. This approach directs the model toward producing more accurate summaries.

\textbf{KL Divergence Loss:} Our experiments reveal that models often generate summaries applicable to several aspects, likely to reduce the average loss across aspects. This propensity could result in the creation of generic, non-aspect-specific summaries, conflicting with our goal of dynamic aspect-based summarization. To counter this, we incorporate KL divergence loss between single-aspect summaries. 

\textbf{Aspect Number Prediction:} Our trials show that a mismatch in aspect numbers between the reference and generated summaries can significantly affect the model's performance. To minimize this difference, we incorporate an objective to predict the number of aspects.

By integrating these objectives, we aim to build a multi-objective framework capable of producing aspect-based summaries that are both accurate and unique to each aspect while accurately predicting the number of aspects.

 \subsection{Loss Weight and Limit Function}\label{sct:weight_and_limitfunc}
Selecting the weights (\(\lambda_1\), \(\lambda_2\), and \(\lambda_3\)) for each objective in the multi-objective learning framework is crucial to balance each task's contribution to the model's final performance. In our method, \(\lambda_1\) is always set to 1, while the values for \(\lambda_2\) and \(\lambda_3\) are determined via grid search. For this process, we set aside a fraction (\(20\% \)) of the original dataset and fine-tuned the model using different combinations of weights. We choose the weight values of \(\lambda_{2}\) and \(\lambda_{3}\) from 0 (which disables the corresponding objective), 0.1, 0.5, or 1. The combination that yielded the best performance on the validation data was chosen as the optimal one for the main experiments.

An alternative approach, which could potentially improve the model's performance, involves adding a layer to the model to learn the optimal weight settings during the fine-tuning process using the complete dataset.

Regarding the limit function (L), its purpose is to ensure numerical stability during fine-tuning. The KL divergence, which measures the disparity between two probability distributions, could potentially become very large or even infinite, destabilizing the fine-tuning process. To prevent this, we employed the limit function to cap the KL divergence to bound the limits (using either the Tanh or Sigmoid functions). However, our experiments indicated that this limit might reduce our model's performance in cases where there was no numerical stability issue.
 
\section{Human Annotators}\label{sct:annotators}
        For this task, each sample is evaluated with three evaluators. The evaluators are paid about 10--15 dollars per hour for this task. We look for evaluators with a bachelor's degree or higher living in the US. We also require that the evaluators have a good command of English. The evaluators are told that this will be used for academic usage.

        We follow the instructions used by~\citet{Guo2024DDABS} for the human evaluation process. We provide the scores of baselines to align the score of the baselines and our \Method{}.

\section{Experimental Details}
\subsection{Computing Infrastructure}\label{sct:computing_infrastructure}
For our experiments, we used a Lambda machine equipped with 250 GB of memory, 4 RTX 8000 GPUs, and 32 CPU cores. The machine runs on Ubuntu 20.04, and our experiments were conducted using Python 3.8.10. The CUDA version is 11.9, and the GPU Driver Version is 520.61. The main packages we utilize include bertopic (0.14.1), cuml-cu11 (23.4.1), deepspeed (0.8.0), torch (1.13.1), scikit-learn (1.1.2), sentence-transformers (2.2.2), scipy (1.9.1), transformers (4.22.1), and numpy (1.23.3). The complete list of packages will be provided in the code release.

\subsection{GPT-3.5-Turbo Details}\label{sct:GPT-3.5-turbo-details}
    In our experiments, we leveraged multiple versions of GPT-3.5-Turbo, adapting to the evolving OpenAI APIs and model updates. The specifics of our model usage are as follows:
    \begin{itemize}
        \item For \DCnnDM{} and \DWikiHow{}, the dataset paper by \citet{Guo2024DDABS} reports zero-shot learning results using GPT-3.5-Turbo-16k-0613. Our few-shot fine-tuning experiments were conducted with GPT-3.5-Turbo-1106.
        \item In the case of \OASUM{}, we employed GPT-3.5-Turbo-1106 for both zero-shot and few-shot learning experiments.
    \end{itemize}

    The determination of the optimal number of samples for fine-tuning was a pivotal aspect of our methodology. Consistent with the insights provided by \citet{Guo2024DDABS}, our observations confirmed that increasing the number of fine-tuning samples does not necessarily lead to a linear improvement in model performance. Consequently, we relied on evaluating the loss values from validation sets obtained during the fine-tuning process to identify the most effective model configuration for the \DCnnDM{} and \OASUM{} datasets. For the \DWikiHow{} dataset, we adopted the fine-tuning settings as recommended by \citet{Guo2024DDABS}. Table~\ref{tab:fine-tuning_performance} presents the observed loss values for different fine-tuning sample sizes for both \DCnnDM{} and \OASUM{}. The process of selecting the sample number was halted once we identified a point of diminishing returns, where additional samples no longer contributed to performance gains. The model configuration that achieved the highest level of performance was subsequently chosen for detailed presentation in the main paper.

    \begin{table}[!hbt]
        \centering
        \begin{tabular}{lll}
        \toprule
            & DCnnDM & OASUM  \\ \midrule
        50  & 1.9651 & 1.4669 \\
        100 & 1.0328 & \textbf{1.2620} \\
        200 & \textbf{0.2759} & 1.3198 \\
        400 & 1.5383 & N/A    \\ \midrule
        \end{tabular}
        \caption{The loss of both datasets on the sampled validation sets with various numbers of fine-tuning samples. The ones with the best performance are bolded.}
        \label{tab:fine-tuning_performance}
    \end{table}

\subsection{Data Processing}\label{sct:Data_process}
    Table~\ref{tab:process_info} outlines the established thresholds for the length of source articles, the length of single-aspect summaries, and the maximum number of aspects considered for truncation during our experiments. Specifically, the threshold for summary length pertains to the length allocated for a single aspect, with the aggregate length of multi-aspect summaries being the product of this threshold and the number of aspects. Our data processing approach aligns with the methodologies employed by \citet{Guo2024DDABS} and \citet{yang-etal-2023-oasum} for baseline comparisons.
    
    For all datasets, truncation of source articles and single-aspect summaries is guided by statistical measures, specifically the mean and standard deviation of their lengths, setting the threshold to approximately the mean plus twice the standard deviation (\(\approx\) mean + 2\(\times\)std). This criterion ensures a balanced representation of content while managing outlier lengths effectively.
    
    In determining the maximum number of aspects (\#Asp) for our datasets, we adopted dataset-specific thresholds to reflect the inherent variability in aspect counts. For \DCnnDM{}, the threshold was set based on the maximum aspect count observed within the dataset, which stands at 11. For the other datasets, recognizing their long-tail distribution in aspect counts, we established thresholds of 16 for \DWikiHow{} and 8 for \OASUM{}. These thresholds approximate the mean plus twice the standard deviation (\(\approx\) mean + 2\(\times\)std), enabling us to encompass a broad spectrum of aspect distributions effectively. Accordingly, samples that exceeded these thresholds were omitted from our study. Additionally, we excluded samples comprising solely of single aspects from \DWikiHow{} and \OASUM{} to prevent single-aspect samples from predominating our experiments, aligning with the approach suggested by \citet{Guo2024DDABS}. This decision ensures a balanced representation across varying aspect counts, enhancing the robustness of our analysis.

    \begin{table}[!hbt]
        \centering
        \setlength\tabcolsep{2.5pt}
        \begin{tabular}{llll}
        \toprule
                  & Arc. Leng & Sum. Length & \#Asp \\ \midrule
        \DCnnDM{}   & 11,264    & 76          & 12   \\
        \DWikiHow{} & 2,040     & 20          & 16   \\
        \OASUM{} & 8192    & 128          & 8    \\ \bottomrule
        \end{tabular}
        \caption{The thresholds for the source article length, single-aspect summary length, and the maximum aspect number for all datasets.}\label{tab:process_info}
    \end{table}

\subsection{BERTopic Hyperparameter Tuning}\label{sct:BERTopic_tuning}
    For the baselines, we performed a grid search for tuning the BERTopic model hyperparameters. 
    The hyperparameters tuned for the BERTopic include ``n\_neighbours'', ``n\_component'', and ``min\_dist'' which control the cluster size and the samples within a cluster. We performed BERTopic clustering on each sample of the validation data and selected the combination of hyperparameters that minimized the absolute difference between the generated and reference aspect numbers.

\subsection{Hyperparameters and Random Seeds}\label{sct:hyper-and-seed}
    For our experiments, we used three random seeds (0, 10, and 42) for the complete dataset experiments. For the ablation analysis and loss weight search, we used only one random seed (42). We used the Hugging Face implementation for fine-tuning the language model with a batch size of 4 due to GPU memory limitations. The training epoch was set to 10 with an early stop, and all other training process hyperparameters were set to the default values provided by the package. We also used DeepSpeed to reduce memory requirements during the fine-tuning process. The specific DeepSpeed configuration will be provided along with our code.

%% file: main.bbl
\begin{thebibliography}{35}
\expandafter\ifx\csname natexlab\endcsname\relax\def\natexlab#1{#1}\fi

\bibitem[{Ahuja et~al.(2022)Ahuja, Xu, Gupta, Horecka, and Durrett}]{ahuja2022aspectnews}
Ojas Ahuja, Jiacheng Xu, Akshay Gupta, Kevin Horecka, and Greg Durrett. 2022.
\newblock Aspectnews: Aspect-oriented summarization of news documents.
\newblock In \emph{Proceedings of the 60th Annual Meeting of the Association for Computational Linguistics (Volume 1: Long Papers)}, pages 6494--6506.

\bibitem[{Amar et~al.(2023)Amar, Schiff, Ernst, Shefer, Shapira, and Dagan}]{amar-etal-2023-openasp}
Shmuel Amar, Liat Schiff, Ori Ernst, Asi Shefer, Ori Shapira, and Ido Dagan. 2023.
\newblock \href {https://aclanthology.org/2023.emnlp-main.121} {{O}pen{A}sp: A benchmark for multi-document open aspect-based summarization}.
\newblock In \emph{Proceedings of the 2023 Conference on Empirical Methods in Natural Language Processing}, pages 1967--1991, Singapore. Association for Computational Linguistics.

\bibitem[{Amplayo et~al.(2021)Amplayo, Angelidis, and Lapata}]{amplayo2021aspect}
Reinald~Kim Amplayo, Stefanos Angelidis, and Mirella Lapata. 2021.
\newblock Aspect-controllable opinion summarization.
\newblock In \emph{Proceedings of the 2021 Conference on Empirical Methods in Natural Language Processing}, pages 6578--6593.

\bibitem[{Angelidis et~al.(2021)Angelidis, Amplayo, Suhara, Wang, and Lapata}]{angelidis2021extractive}
Stefanos Angelidis, Reinald~Kim Amplayo, Yoshihiko Suhara, Xiaolan Wang, and Mirella Lapata. 2021.
\newblock Extractive opinion summarization in quantized transformer spaces.
\newblock \emph{Transactions of the Association for Computational Linguistics}, 9:277--293.

\bibitem[{Angelidis and Lapata(2018)}]{angelidis2018summarizing}
Stefanos Angelidis and Mirella Lapata. 2018.
\newblock Summarizing opinions: Aspect extraction meets sentiment prediction and they are both weakly supervised.
\newblock In \emph{Proceedings of the 2018 Conference on Empirical Methods in Natural Language Processing}, pages 3675--3686.

\bibitem[{Beltagy et~al.(2020)Beltagy, Peters, and Cohan}]{beltagy2020longformer}
Iz~Beltagy, Matthew~E Peters, and Arman Cohan. 2020.
\newblock Longformer: The long-document transformer.
\newblock \emph{arXiv preprint arXiv:2004.05150}.

\bibitem[{Chu and Liu(2019)}]{chu2019meansum}
Eric Chu and Peter Liu. 2019.
\newblock Meansum: A neural model for unsupervised multi-document abstractive summarization.
\newblock In \emph{International Conference on Machine Learning}, pages 1223--1232. PMLR.

\bibitem[{Fabbri et~al.(2021)Fabbri, Kry{\'s}ci{\'n}ski, McCann, Xiong, Socher, and Radev}]{fabbri2021summeval}
Alexander~R Fabbri, Wojciech Kry{\'s}ci{\'n}ski, Bryan McCann, Caiming Xiong, Richard Socher, and Dragomir Radev. 2021.
\newblock Summeval: Re-evaluating summarization evaluation.
\newblock \emph{Transactions of the Association for Computational Linguistics}, 9:391--409.

\bibitem[{Fan et~al.(2018)Fan, Grangier, and Auli}]{fan2018controllable}
Angela Fan, David Grangier, and Michael Auli. 2018.
\newblock Controllable abstractive summarization.
\newblock In \emph{Proceedings of the 2nd Workshop on Neural Machine Translation and Generation}, pages 45--54.

\bibitem[{Ficler and Goldberg(2017)}]{ficler2017controlling}
Jessica Ficler and Yoav Goldberg. 2017.
\newblock Controlling linguistic style aspects in neural language generation.
\newblock In \emph{Proceedings of the Workshop on Stylistic Variation}, pages 94--104.

\bibitem[{Frermann and Klementiev(2019)}]{frermann2019inducing}
Lea Frermann and Alexandre Klementiev. 2019.
\newblock Inducing document structure for aspect-based summarization.
\newblock In \emph{Proceedings of the 57th Annual Meeting of the Association for Computational Linguistics}, pages 6263--6273.

\bibitem[{Goyal et~al.(2022)Goyal, Li, and Durrett}]{goyal2022news}
Tanya Goyal, Junyi~Jessy Li, and Greg Durrett. 2022.
\newblock News summarization and evaluation in the era of gpt-3.
\newblock \emph{arXiv preprint arXiv:2209.12356}.

\bibitem[{Grootendorst(2022)}]{grootendorst2022bertopic}
Maarten Grootendorst. 2022.
\newblock Bertopic: Neural topic modeling with a class-based tf-idf procedure.
\newblock \emph{arXiv preprint arXiv:2203.05794}.

\bibitem[{Guo and Vosoughi(2023)}]{guo-vosoughi-2023-length}
Xiaobo Guo and Soroush Vosoughi. 2023.
\newblock \href {https://doi.org/10.18653/v1/2023.emnlp-main.984} {Length does matter: Summary length can bias summarization metrics}.
\newblock In \emph{Proceedings of the 2023 Conference on Empirical Methods in Natural Language Processing}, pages 15869--15879, Singapore. Association for Computational Linguistics.

\bibitem[{Guo and Vosoughi(2024)}]{Guo2024DDABS}
Xiaobo Guo and Soroush Vosoughi. 2024.
\newblock Disordered-dabs: A benchmark for dynamic aspect-based summarization in disordered texts.
\newblock \emph{arXiv preprint arXiv:2402.10554}.

\bibitem[{Hayashi et~al.(2021)Hayashi, Budania, Wang, Ackerson, Neervannan, and Neubig}]{hayashi2021wikiasp}
Hiroaki Hayashi, Prashant Budania, Peng Wang, Chris Ackerson, Raj Neervannan, and Graham Neubig. 2021.
\newblock Wikiasp: A dataset for multi-domain aspect-based summarization.
\newblock \emph{Transactions of the Association for Computational Linguistics}, 9:211--225.

\bibitem[{He et~al.(2022)He, Kryscinski, McCann, Rajani, and Xiong}]{he-etal-2022-ctrlsum}
Junxian He, Wojciech Kryscinski, Bryan McCann, Nazneen Rajani, and Caiming Xiong. 2022.
\newblock \href {https://aclanthology.org/2022.emnlp-main.396} {{CTRL}sum: Towards generic controllable text summarization}.
\newblock In \emph{Proceedings of the 2022 Conference on Empirical Methods in Natural Language Processing}, pages 5879--5915, Abu Dhabi, United Arab Emirates. Association for Computational Linguistics.

\bibitem[{Koupaee and Wang(2018)}]{koupaee2018wikihow}
Mahnaz Koupaee and William~Yang Wang. 2018.
\newblock Wikihow: A large scale text summarization dataset.
\newblock \emph{arXiv preprint arXiv:1810.09305}.

\bibitem[{Liu et~al.(2018)Liu, Saleh, Pot, Goodrich, Sepassi, Kaiser, and Shazeer}]{liugenerating}
Peter~J Liu, Mohammad Saleh, Etienne Pot, Ben Goodrich, Ryan Sepassi, Lukasz Kaiser, and Noam Shazeer. 2018.
\newblock Generating wikipedia by summarizing long sequences.
\newblock In \emph{International Conference on Learning Representations}.

\bibitem[{Lu et~al.(2009)Lu, Zhai, and Sundaresan}]{lu2009rated}
Yue Lu, ChengXiang Zhai, and Neel Sundaresan. 2009.
\newblock Rated aspect summarization of short comments.
\newblock In \emph{Proceedings of the 18th international conference on World wide web}, pages 131--140.

\bibitem[{Maddela et~al.(2022)Maddela, Kulkarni, and Preo{\c{t}}iuc-Pietro}]{maddela2022entsum}
Mounica Maddela, Mayank Kulkarni, and Daniel Preo{\c{t}}iuc-Pietro. 2022.
\newblock Entsum: A data set for entity-centric extractive summarization.
\newblock In \emph{Proceedings of the 60th Annual Meeting of the Association for Computational Linguistics (Volume 1: Long Papers)}, pages 3355--3366.

\bibitem[{Peng et~al.(2023)Peng, Alcaide, Anthony, Albalak, Arcadinho, Cao, Cheng, Chung, Grella, GV et~al.}]{peng2023rwkv}
Bo~Peng, Eric Alcaide, Quentin Anthony, Alon Albalak, Samuel Arcadinho, Huanqi Cao, Xin Cheng, Michael Chung, Matteo Grella, Kranthi~Kiran GV, et~al. 2023.
\newblock Rwkv: Reinventing rnns for the transformer era.
\newblock \emph{arXiv preprint arXiv:2305.13048}.

\bibitem[{Poli et~al.(2023)Poli, Massaroli, Nguyen, Fu, Dao, Baccus, Bengio, Ermon, and R{\'e}}]{poli2023hyena}
Michael Poli, Stefano Massaroli, Eric Nguyen, Daniel~Y Fu, Tri Dao, Stephen Baccus, Yoshua Bengio, Stefano Ermon, and Christopher R{\'e}. 2023.
\newblock Hyena hierarchy: Towards larger convolutional language models.
\newblock \emph{arXiv preprint arXiv:2302.10866}.

\bibitem[{See et~al.(2017)See, Liu, and Manning}]{see2017get}
Abigail See, Peter~J Liu, and Christopher~D Manning. 2017.
\newblock Get to the point: Summarization with pointer-generator networks.
\newblock In \emph{Proceedings of the 55th Annual Meeting of the Association for Computational Linguistics (Volume 1: Long Papers)}, pages 1073--1083.

\bibitem[{Sennrich et~al.(2016)Sennrich, Haddow, and Birch}]{sennrich2016controlling}
Rico Sennrich, Barry Haddow, and Alexandra Birch. 2016.
\newblock Controlling politeness in neural machine translation via side constraints.
\newblock In \emph{Proceedings of the 2016 Conference of the North American Chapter of the Association for Computational Linguistics: Human Language Technologies}, pages 35--40.

\bibitem[{Souza and Manzato(2022)}]{souza2022aspect}
Luan Soares~de Souza and Marcelo~Garcia Manzato. 2022.
\newblock Aspect-based summarization: an approach with different levels of details to explain recommendations.
\newblock In \emph{Proceedings of the Brazilian Symposium on Multimedia and the Web}, pages 202--210.

\bibitem[{Tan et~al.(2020)Tan, Qin, Xing, and Hu}]{tan2020summarizing}
Bowen Tan, Lianhui Qin, Eric Xing, and Zhiting Hu. 2020.
\newblock Summarizing text on any aspects: A knowledge-informed weakly-supervised approach.
\newblock In \emph{Proceedings of the 2020 Conference on Empirical Methods in Natural Language Processing (EMNLP)}, pages 6301--6309.

\bibitem[{Wang et~al.(2022)Wang, Pang, Chen, Phang, and Bowman}]{wang2022squality}
Alex Wang, Richard~Yuanzhe Pang, Angelica Chen, Jason Phang, and Samuel Bowman. 2022.
\newblock Squality: Building a long-document summarization dataset the hard way.
\newblock In \emph{Proceedings of the 2022 Conference on Empirical Methods in Natural Language Processing}, pages 1139--1156.

\bibitem[{Wang and Ling(2016)}]{wang2016neural}
Lu~Wang and Wang Ling. 2016.
\newblock Neural network-based abstract generation for opinions and arguments.
\newblock In \emph{Proceedings of NAACL-HLT}, pages 47--57.

\bibitem[{Yang et~al.(2018)Yang, Qu, Shen, Liu, Zhao, and Zhu}]{yang2018aspect}
Min Yang, Qiang Qu, Ying Shen, Qiao Liu, Wei Zhao, and Jia Zhu. 2018.
\newblock Aspect and sentiment aware abstractive review summarization.
\newblock In \emph{Proceedings of the 27th international conference on computational linguistics}, pages 1110--1120.

\bibitem[{Yang et~al.(2023{\natexlab{a}})Yang, Li, Zhang, Chen, and Cheng}]{yang2023exploring}
Xianjun Yang, Yan Li, Xinlu Zhang, Haifeng Chen, and Wei Cheng. 2023{\natexlab{a}}.
\newblock Exploring the limits of chatgpt for query or aspect-based text summarization.
\newblock \emph{arXiv preprint arXiv:2302.08081}.

\bibitem[{Yang et~al.(2023{\natexlab{b}})Yang, Song, Cho, Wang, Pan, Petzold, and Yu}]{yang-etal-2023-oasum}
Xianjun Yang, Kaiqiang Song, Sangwoo Cho, Xiaoyang Wang, Xiaoman Pan, Linda Petzold, and Dong Yu. 2023{\natexlab{b}}.
\newblock \href {https://doi.org/10.18653/v1/2023.findings-acl.268} {{OAS}um: Large-scale open domain aspect-based summarization}.
\newblock In \emph{Findings of the Association for Computational Linguistics: ACL 2023}, pages 4381--4401, Toronto, Canada. Association for Computational Linguistics.

\bibitem[{Zhong et~al.(2021)Zhong, Yin, Yu, Zaidi, Mutuma, Jha, Hassan, Celikyilmaz, Liu, Qiu et~al.}]{zhong2021qmsum}
Ming Zhong, Da~Yin, Tao Yu, Ahmad Zaidi, Mutethia Mutuma, Rahul Jha, Ahmed Hassan, Asli Celikyilmaz, Yang Liu, Xipeng Qiu, et~al. 2021.
\newblock Qmsum: A new benchmark for query-based multi-domain meeting summarization.
\newblock In \emph{Proceedings of the 2021 Conference of the North American Chapter of the Association for Computational Linguistics: Human Language Technologies}, pages 5905--5921.

\bibitem[{Zhu et~al.(2021)Zhu, Tu, Shi, Li, Hou, and Cui}]{zhu2021twag}
Fangwei Zhu, Shangqing Tu, Jiaxin Shi, Juanzi Li, Lei Hou, and Tong Cui. 2021.
\newblock Twag: A topic-guided wikipedia abstract generator.
\newblock In \emph{Proceedings of the 59th Annual Meeting of the Association for Computational Linguistics and the 11th International Joint Conference on Natural Language Processing (Volume 1: Long Papers)}, pages 4623--4635.

\bibitem[{Zhu et~al.(2022)Zhu, Dong, Wei, Qin, and Liu}]{zhu2022transforming}
Haichao Zhu, Li~Dong, Furu Wei, Bing Qin, and Ting Liu. 2022.
\newblock Transforming wikipedia into augmented data for query-focused summarization.
\newblock \emph{IEEE/ACM Transactions on Audio, Speech, and Language Processing}, 30:2357--2367.

\end{thebibliography}
